\UseRawInputEncoding
\def\year{2019}\relax
\documentclass[letterpaper]{article} %DO NOT CHANGE THIS
\usepackage{aaai19}  %Required
\usepackage{times}  %Required
\usepackage{helvet}  %Required
\usepackage{courier}  %Required
\usepackage{url}  %Required
\usepackage{graphicx}  %Required
\usepackage{epsfig}
\usepackage{subfigure}
\usepackage{amsmath}
\usepackage{amssymb}
\usepackage{caption}
\usepackage[ruled,vlined]{algorithm2e}
\usepackage{lineno}
\usepackage{booktabs}
\usepackage{color}
\usepackage{multirow,booktabs}
\usepackage{wasysym}
\usepackage{array}

\begin{document}
% The file aaai.sty is the style file for AAAI Press
% proceedings, working notes, and technical reports.
%
\title{Cooperative Multi-Agent Policy Gradients with Sub-optimal Demonstration}
\author{Peixi Peng$^1$,~~~~~Junliang Xing$^1$\\
1, Institute of Automation, Chinese Academy of Sciences
}
\maketitle
\begin{abstract}
Many reality tasks such as robot coordination can be naturally modelled as multi-agent cooperative system where the rewards are sparse.
This paper focuses on learning  decentralized policies  for such tasks using  sub-optimal demonstration. To learn the multi-agent cooperation effectively and tackle the sub-optimality of demonstration, a self-improving learning method is proposed: On the one hand, the centralized state-action values are initialized by the demonstration and updated by the learned decentralized policy to improve the sub-optimality. On the other hand,  the Nash Equilibrium  are found by the current state-action value and are used as a guide to learn the policy. The proposed method is evaluated on the combat RTS games which requires  a high level of multi-agent cooperation.  Extensive experimental results on various combat scenarios demonstrate that the proposed method can learn multi-agent cooperation effectively. It significantly outperforms many state-of-the-art demonstration based approaches.

\end{abstract}
\section{Introduction}
By combining with deep learning models \cite{lecun2015deep}, reinforcement learning (RL) has achieved successful applications to many  challenging problems in these years such as Go \cite{Silver2017Mastering}, general Atari
game-playing \cite{mnih2015human} and robot motor control \cite{Levine2016End}. Most of these works involve only a single agent. However, many real-world problems such as autonomous vehicle coordination \cite{Cao2012An}, network packet delivery \cite{Wang2005Multi} and playing combat games \cite{usunier2016episodic} are naturally modelled as cooperative multi-agent systems \cite{Stone2000Multiagent}. The aim is coordinating multiple agents to achieve a unified goal or maximum team benefits by cooperation.

Since the joint state-action space of the agents grows exponentially with the number of agents, RL methods designed for single agents typically fare poorly on such tasks.
To cope with this complexity, it is often necessary to resort to decentralised policies, in which each agent selects its own action conditioned only on its local action-observation history \cite{foerster2017counterfactual}. RL entails the knowledge of the reward function, or at least observations of immediate reward. Unfortunately, the rewards of multi-agent states need to take into account a wide range of factors such as the joint states of all agents. Hence, it is very difficult to calculate the reward of most states in many reality multi-agent systems. By contrast, it is often quite natural to express a task goal as a sparse reward function \cite{Ve2017Leveraging}. For example, the terminal state reward of a combat game can be easily defined by victory or defeat, while it is hard to evaluate the combat situation at non-terminal states.  Another typical example is the multiple robot coordination \cite{chen2017decentralized,Everett2018Motion} where the rewards are only observed directly in a few situations such as collision or arrival at the destination.
A widely-used approach to the sparse rewards is learning from demonstrations (LfD) \cite{schaal1997learning,Hester2017Deep}. In many reality applications, although there is very little or no knowledge of the reward function, but there is  access to demonstrations performed by experts in the task. A natural method is to recover experts' strategies from available demonstrations by supervised learning \cite{Ross2010A}. It assumes that the experts are performing well. Agents should also be able to perform well by simply mimicking experts' moves. Another type of approaches, termed inverse reinforcement learning (IRL) \cite{Ng2000Algorithms}, formulates the task as solving
an inverse problem, and strives to infer the hidden reward function from expert demonstrations. Most existing IRL methods assume that the demonstrations are generated by an optimal strategy.
However, due to the extremely large state action space of multi-agents systems, it's hard to design an optimal strategy manually even for experts. The available demonstrations are always sub-optimal, and  need to be improved rather than just imitating the demonstrator or inferring the reward from the demonstration directly.

Another crucial challenge is multi-agent credit assignment \cite{NIPS2003_2476}:  In cooperative setting, the agents�� actions are performed simultaneously and the state-action value are centralized and shared by all agents. Hence, even if the centralized state-action value can be evaluated from a demonstration, it is still hard to identify  each agent action's own contribution. Some existing works design individual reward functions for each agent by one agent action exploration \cite{tampuu2017multiagent}. However, these rewards ignore multi-agent cooperation and often fail to encourage individual agents to sacrifice themselves for team advantages. For example, in a combat, the healthy agents should go forward to draw enemy fire and cover any injured allied agents. This task is always failed when each agent is only driven by its individual reward. Alternative type of methods are designed by centralised learning \cite{sukhbaatar2016learning}. However, it is inefficient to explore all agents' actions simultaneously, because the joint state-action space of multiple agents is extremely large.
\begin{figure}[tp]
	\centering
	\setlength{\belowcaptionskip}{0pt}
	\includegraphics[width = 0.9\linewidth]{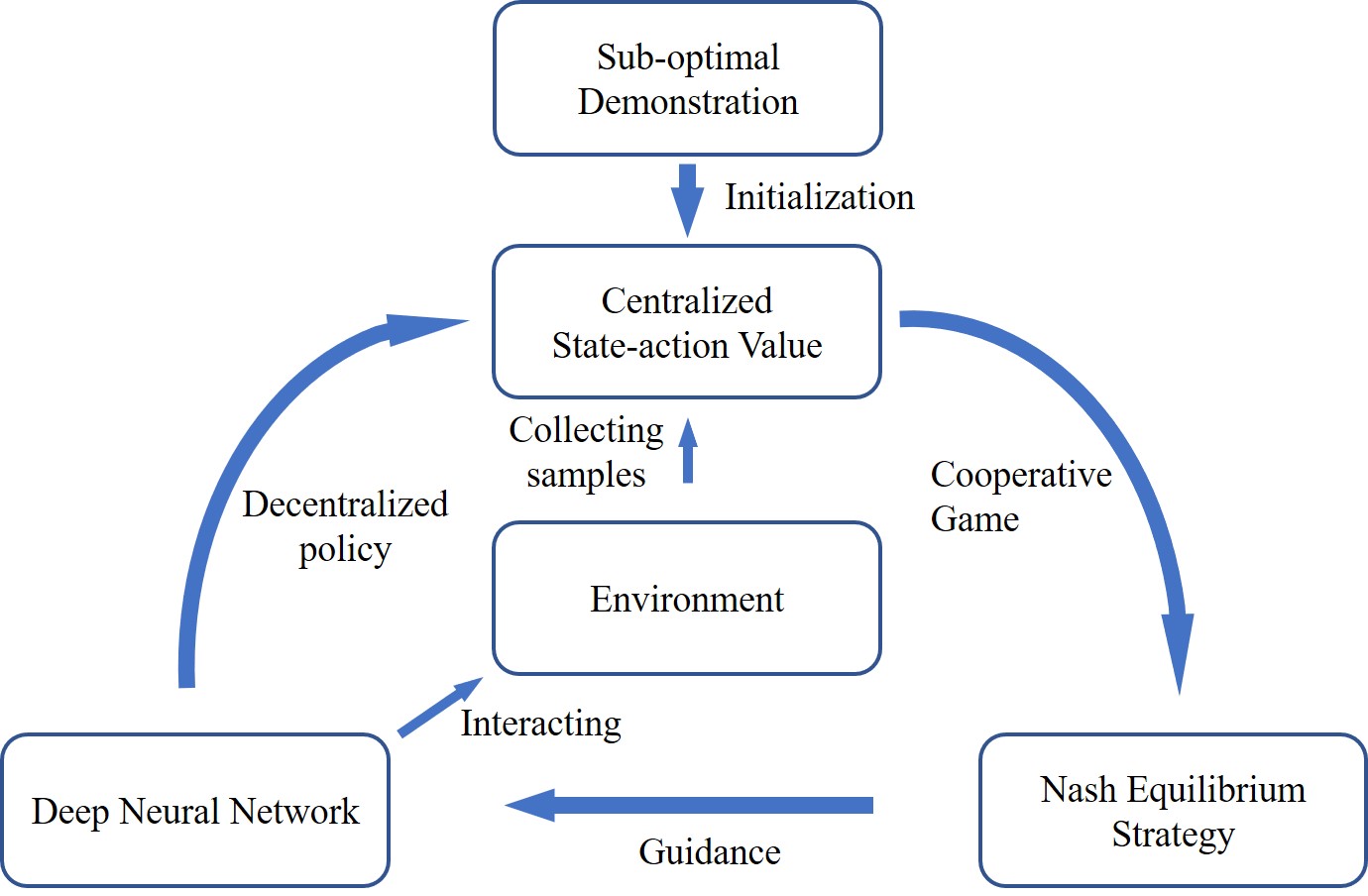}
	\caption{The framework of the proposed method.}
  \label{fig:framework}
\end{figure}

To cope with these difficulties, a novel RL method is proposed to learn multi-agent cooperation by sub-optimal demonstrations. As shown in Fig.~\ref{fig:framework}, the proposed method involves self-improvement: On the one hand, although the demonstrations are not optimal, it still can be assumed that experts perform well. Hence, the state-action value can be initialized by the experts' demonstrations at the early stage of learning. To take the sub-optimality of demonstration into account, the state-action value is updated by the learned policy during the learning procedure. In other words, the sub-optimal demonstrations offer  initial guidance to ensure that the learned policy is better than experts.
On the other hand, the tasks are modeled as the cooperative game and  the Nash Equilibrium strategies are found  by current centralized state-action value. These Nash Equilibrium strategies can be used to learn decentralised policies and  it is the main difference between the proposed method with other multi-agent policy methods \cite{foerster2017counterfactual,sukhbaatar2016learning}.
The Nash Equilibrium reduces the state-action exploration space and  ensures more effective learning of multi-agent cooperation.
These two steps can complement and improve each other: the centralized state-action value can be used to find the Nash Equilibrium strategies to learn better decentralized policy; whilst the learned decentralized polices can help to calculate more accurate state-action values.

To summarize, the main contributions of this work are threefold: (1) a novel learning algorithm is proposed to learn multi-agent cooperation with sparse reward by sub-optimal demonstration, (2) the task is modeled as cooperative game and  Nash Equilibrium strategy is found to learn decentralized policy, and (3) the state-action value function is updated in learning to tackle the sub-optimality of demonstration.
The rest of the paper is organized as follows: Section 2 reviews the related work and highlights our technical contributions. Section 3 formulates the multi-agent learning task by the cooperative game. The learning algorithm is described in Section 4. Extensive experimental analysis and comparisons are discussed in Section 5. The paper concludes with Section 6.

\section{Related Work}
\label{sec_relwork}
\textbf{Multi-agent Reinforcement Learning (MARL)} methods have historically been applied in many settings \cite{Tuyls2012Multiagent,Busoniu2008A,Tuyls2012Multiagent}. They have been restricted to tabular methods and simple environments. Motivated from the success of deep RL \cite{mnih2015human} where value/policy is approximated by deep neural networks, the deep MARL methods can scale to high dimensional input and action spaces \cite{tampuu2017multiagent}. Independent Q-learning \cite{Tan1993Multi} is proposed to learn decentralised value functions or policies for each agent independently. It is extended to deep neural networks  in \cite{tampuu2017multiagent} by combining with DQN  \cite{mnih2015human}. Some methods \cite{pmlr-v70-foerster17b,pmlr-v70-omidshafiei17a} are proposed to address learning stabilisation following this idea. However, these methods always learn each agent independently and ignore the cooperation among agents.

Another type of methods focuses on centralised learning of joint actions which can naturally handle cooperation problems. However, the joint action space will be extremely large when the system contains a lots of agents,which may lead the learning inefficient.
Coordination graphs \cite{Guestrin2001Multiagent} is proposed to exploit conditional independencies between agents by decomposing a global reward function into a sum of agent-local terms. Sparse cooperative Q-learning \cite{Kok2006Collaborative} proposes a
tabular Q-learning algorithm that learns to coordinate the actions of a group of cooperative agents only in the states where such coordination is necessary.
Recently, more centralised learning methods are proposed to multi-agent communication by combing with deep learning:
CommNet, as designed in \cite{sukhbaatar2016learning}, uses a centralised network architecture
to handle communication between agents. The bidirectionally-coordinated network (BiCNet) is introduced in \cite{peng2017multiagent} to maintain a scalable and effective communication protocol among multiple units, and it additionally requires estimating individual agent
rewards. However, only centralized state-value is known in our cooperative setting.
GMEZO \cite{usunier2016episodic} models the multi-agent system as a greedy Markov decision procedure and present an RL algorithm using first-order optimisation. To handle an extremely large number of agents, mean field RL \cite{pmlr-v80-yang18d} is proposed to describe the interaction of an agent with its neighboring agents.
Recently, several works are designed as hybrid approaches. QMIX \cite{pmlr-v80-rashid18a} decomposes the centralized state-action values as a group of single agent's values and enforce that $\arg\max$ performed on the centralized value yields the same result as
a set of individual $\arg\max$ operations performed on each agent. This assumption fails in some cooperation tasks where an agent should sacrifice itself to the benefit of the team. In addition, several methods are designed by learning decentralized policy using centralized critic.
COMA \cite{foerster2017counterfactual} estimates a counterfactual advantage function to update decentralized policy.
Similarly, \cite{NIPS2017_7217} learn a centralised critic for each agent and apply this to
competitive games with continuous action spaces. In these methods, one agent's policy is updated by assuming other agents' policies is known as on-policy.
Hence, they have poor sample efficiency and are prone to getting stuck in sub-optimal local minima \cite{pmlr-v80-rashid18a}.
Different with them, the proposed method solves the the Nash Equilibrium to learn decentralised policies.  The Nash Equilibrium reduces the exploration state-action space and leads to more effective learning of cooperation. A similar method is proposed in \cite{NIPS2017_7007} where one agent update it's policy by sampling other agent's policies from their individual meta-strategies. Compared with \cite{NIPS2017_7007}, the proposed method assumes all agents are performing as Nash Equilibrium jointly and it is more reasonable to multi-agent cooperation. All of these deep MARL methods assume that the reward is available at every state.

\textbf{Learning from Demonstration (LfD)} \cite{schaal1997learning} has received increasing attention as a promising way to tackle  the problem of sparse rewards. Imitation learning is primarily concerned with recovering the strategy of the demonstrator in a supervised fashion. DAGGER \cite{Ross2010A} formulates the task as essentially a regression problem and sets the objective of minimizing the prediction error. Deeply AggreVaTeD \cite{Sun2017Deeply} extends DAGGER to continuous action spaces by combining with deep neural networks. These methods just imitate the demonstration and can not improve upon the expert. To improve the demonstration, several methods \cite{mnih2015human,OGTL} train the network by combining imitation learning and RL together. However, AlphaGo \cite{mnih2015human} is designed for a single agent and OGTL \cite{OGTL} ignores multi-agent cooperation during learning. Recently, a few approaches are proposed to explore the sparse-reward environment by LfD. For instance, DQfD \cite{Hester2017Deep} introduces LfD into DQN adding demonstration data into the replay buffer. Similarly, DDPGfD \cite{Ve2017Leveraging} extends LfD to continuous action domains. POfD \cite{pmlr-v80-kang18a} leverage demonstrations to guide exploration through enforcing occupancy measure matching between the learned policy and current demonstrations. These methods are all designed for single agent systems and aim at exploring around demonstration to find a better policy. However, the joint state-action space of multiple agents is very large and the optimal policy may be far from experts' policy with a large probability. Another type of LfD methods is inverse reinforcement learning (IRL) \cite{Ng2000Algorithms} which targets at recovering the reward function of a given task from samples, and it is extended to two-player zero-sum game \cite{Syed2008Apprenticeship,Syed2008A} by alternating between fitting the reward function
and selecting the policy.  GAIL \cite{Ho2016Generative} applies the IRL to high-dimensional continuous
control problems. The multi-agent IRL methods \cite{Reddy2012Inverse,Lin2018Multiagent} are proposed to a non-cooperative game and  zero-sum
stochastic games respectively. All these methods assume that the  demonstration is optimal. Besides these methods,   \cite{pmlr-v80-wang18d} propose to handle sub-optimal demonstrations in two-player zero-sum competitive games. The centralized rewards are inferred by two-player zero-sum constraint. A  network is utilized to approximate the rewards by centralized learning. However, the proposed method focuses on learning decentralized policy in multi-agent cooperative tasks which don't satisfy the two-player zero-sum constraint.

\section{Problem Definition}
\label{sec_model}
In the Markov Decision Procedure (MDP) model, the multi-agent system with sparse reward has the following components:
\begin{itemize}
  \item \textbf{States:} The states are modeled by a time series $S=\{s_0,s_1,...,s_T\}$ where $s_t$ stands for the state at time $t$ and $s_T$ is the terminal state.
  \item \textbf{Agents:} $G=\{g_1,g_2,...,g_{N}\}$ are agents and $N$ is the number of the agents. Note the proposed method can be still applicative when $N$ is varied at different states.
  \item \textbf{Actions:} $\{A^g(s)|s\in S\}$ defines the action space for agent $g\in G$. $A^g(s)$ is the finite set of candidate action set that can be performed by agent $g$ at state $s$. For simplicity, suppose that the candidate actions set of agent $g$ is same at every state, then $A^g$ refers to the available action set for agent $g$. At each time step $t$,  each agent $g$ simultaneously chooses an action $a^g_t \in A^g$, forming a joint action set $A_t=\{a^g_t\}_{g \in G}$.
  \item \textbf{State transition:} $\mathbb{P}_{t+1}=\mathbb{P}(s_{t+1}|s_t,A_t)$ is the state transition probability. In an MDP, the transition probability depends only on the current state and the agents' actions at the corresponding state.
  \item \textbf{Decentralized policy:}  Given a state $s$ and an agent $g\in G$, the decentralized policy is the set of probability distributions over all actions: $p(a|g,s)$ where $a\in A^g$. The decentralized policy learning is to find a optimal policy $\{p_{\theta}(a|g,s)\}_{a \in A^g}$ where $\theta$ is the model parameters (i.e, the network parameters in the proposed method).
  \item \textbf{Rewards:}  The reward function $R(s_{t+1}| s_t, A_t)$ is bounded and used to measure the reward (or payoff) of the states transition: $s_t \times A_t \times s_{t+1} \rightarrow  R$. We use $R(s_{t+1})$ to represent it for simplicity. In our task, rewards can only be defined in a few states. Without loss of generality, the rewards of the terminal states can be calculated as $R(s_{T})$ and  $R(s_{t})=0$ at  other states.
  \item \textbf{State-action value (Q):} Given a policy $p$ , the $p$ based state-action value $Q^p(s,A)$ can be calculated by the cumulative rewards:
\begin{align}
\small
\begin{aligned}
 \label{eq:ostatevalue}
 Q^p(s,A)&= \mathbb{E}_{\{s_{t+1}\sim \mathbb{P}_{t+1},s_1, A_t\sim p\}_{t=1}^{T-1}}\sum\limits_{t=1}^{T}{\lambda^{t-1}R(s_{t})}\\
 &=\mathbb{E}_{\{s_{t+1}\sim \mathbb{P}_{t+1}, s_1, A_t\sim p\}_{t=1}^{T-1}}{\lambda^{T-1} R(s_{T})},
\end{aligned}
\end{align}
where $s_1 \sim \mathbb{P}(s_1|s,A)$, $\lambda\in [0,1]$ is a discount factor and $\mathbb{E}$ is the expectation.  The state-action value measures best cumulative rewards as $ Q^*(s,A)=\sup_{p}Q^p(s,A)$. Since the multiple agents' actions are performed simultaneously, all the agents share the state-action value.

%  \item \textbf{Cooperative game and pure Nash Equilibrium:} The multi-agent cooperation can be modelled naturally as a cooperative game which focus on how much collective payoff a set of players can gain by forming a coalition. Given the reward $R(s_{t+1})$, the pure Nash Equilibrium (PNE) at state $s_t$ is a set of actions $\tilde A=\{\tilde a^{g_1},...,\tilde a^{g_N}\}$ that if for every unilateral deviation  Since the multiple agents' actions are performed simultaneously, hence the agents all share the same reward function.
%      $a^g\in A^g$:
%      \begin{align}
%    \small
%    \begin{aligned}
%    \label{eq:pne}
%        \mathbb{E}_{s_{t+1}\sim P(s_{t+1}|s_t, a^g, \tilde A^{-g})}R(s_{t+1})\leq  \mathbb{E}_{s_{t+1}\sim P(s_{t+1}|s_t,\tilde A)}R(s_{t+1}),
%    \end{aligned}
%    \end{align}
%    where $\tilde A^{-g}=\prod\limits_{g'\neq g}\{\tilde a^{g'}\}$.
  \item \textbf{Demonstrations:} The demonstrations usually fall into two types: rule-based heuristics and experience data. The former are always designed by expert's prior knowledge. The heuristics are task-specific and the proposed method is independent of  the knowledge of its internal workings. The only requirement is the heuristics can be simulated.
      The later are the observation set $D=\{s_m, a^{g_1}(s_m),...,a^{g^N}(s_m)\}_{m=1}^M$ where shows how experts have performed in the task. Here $s_m$ is the visited state and $\{a^{g_1}(s_m),...,a^{g^N}(s_m)\}$ are each agent's actions  which are performed by experts at the corresponding state $s_m$. Besides,  $m$ stands for the $m$th observation and is irrelevant to the time.
      In order to make the expert's knowledge available at arbitrary states, a parameterized policy  $p_{\theta^D}$ is used to imitate the experts. It can be trained by minimizing the cross-entropy loss function:
       \begin{align}
    \small
    \begin{aligned}
    \label{eq:imtationloss}
   \sum\limits_{m=1}^M{\sum\limits_{g \in G}{-log(p_{\theta^D}(a^{g}(s_m)|^i,s_m)))}}
    \end{aligned}
    \end{align}
    To summarize, these two types of demonstrations can both be written as a policy  $p^D(a|g,s)$. A demonstration is optimal if the multi-agent system can get the highest state-action value according to the demonstration policy at every state, i.e, $Q(s,A)=Q^D(s,A)$ where $Q^D(s,A)$ is a simplification to $Q^{p^D}(s,A)$ . Otherwise, the demonstration is sub-optimal.
\end{itemize}

\section{Learning Algorithm}
\label{sec_Methodology}
%\subsection{State-action Value}
%Given a policy $p(a|g,s)$ for each agent and state, the state value $V^p(s)$ based on $p$ can be calculated by the cumulative rewards:
%\begin{align}
%\small
%\begin{aligned}
% \label{eq:ostatevalue}
% V^p(s)&=\mathbb{E}_{\{s_{t+1}\sim P(s_{t+1}|s_t,A_t), A_t\sim p\}_{t=1}^{T-1}}\sum\limits_{t=1}^{T-1}{\lambda^{t-1}R(s_{t+1}|s_t, A_t)}\\
% &=\mathbb{E}_{\{s_{t+1}\sim P(s_{t+1}|s_t,A_t), A_t\sim p\}_{t=1}^{T-1}}{\lambda^{T-1} R(s_{T}|s_{T-1}, A_{T-1})},
%\end{aligned}
%\end{align}
%where $s_1=s$ and $\lambda\in [0,1]$ is the discount factor. $\mathbb{E}$ is the expectation of $R(s_{t+1})$ according to the states transition probability and the given policy $p$. The state value, which is used to evaluate the state, can be calculated as $ V^*(s)=\sup_{p}V^p(s)$. Based on the state value function and Q-function, the advantage function measuring benefits of actions is defined as state-action value:
%\begin{align}
%\small
%\begin{aligned}
% \label{eq:mactionvalue}
% V(s_t, A_t)=\mathbb{E}_{s_{t+1}\sim P(s_{t+1}|s_t,A_t)}V^*(s_{t+1})-V^*(s_t).
%\end{aligned}
%\end{align}

\subsection{Cooperative Game and Pure Nash Equilibrium}
The multi-agent cooperation can be naturally modelled  as a cooperative game which focuses on how much collective payoff a set of agents can gain by forming a coalition. Given the state-action value $Q(s, A)$, the pure Nash Equilibrium (PNE) at state $s$ is a set of actions $\tilde A=\{\tilde a^{g}\}_{g \in G}$ that if for every unilateral deviation $a^g\in A^g$:
\begin{align}
\small
\begin{aligned}
\label{eq:pne}
Q(s,a^g,\tilde A^{-g}) \leq Q(s, \tilde A)
\end{aligned}
\end{align}
where $\tilde A^{-g}=\prod\limits_{g'\neq g}\{\tilde a^{g'}\}$.
To solve $\tilde A$, the best response dynamic algorithm is utilized in Alg.~\ref{alg:best-response-dynamics}. It can be easily proved that  Alg.~\ref{alg:best-response-dynamics} is convergent to the PNE:  In every iteration, the state reward caused by deviator action  strictly increases. Since the game is finite and the state-action value is bounded, hence best-response dynamics eventually halts, necessarily at a PNE. Alg.~\ref{alg:best-response-dynamics} can be considered as joint state-action space exploration. It approximates the optimal policy from a exponential space ($O(|A|^{|G|})$) in polynomial complexity $O(|A|{|G|})$.

\begin{algorithm}[t]
\small
	\KwIn{The state $s_t$, multiple agents $G=\{g_1,...,g_N\}$ and $Q$ function.}
	\KwOut{The PNE $\tilde A=\{\tilde a^{g_1},...,\tilde a^{g_N}\}$.}
    Initialize the action set $\tilde A \sim p^D(a|g,s)$.\\
    \For{$iter=1,2,...,Iterations$}
    {
        \For{$g \in G$}
        {
            Calculate the response for each action $a$: $Q(s,a,\tilde A^{-g})$.\\
            Choose the action $a^g=\arg\max_{a\in A^g}{Q(s,a,\tilde A^{-g})}$.\\
            Update $\tilde A$ by $\tilde a^g=a^g$.\\
        }
    }
\caption{The best response dynamics algorithm in multi-agent cooperative game.}
\label{alg:best-response-dynamics}
\end{algorithm}

\subsection{Decentralized Policy Gradient}
PNE is a set of discrete actions which may lose useful information. Considering two typical states, the responses (Alg.~\ref{alg:best-response-dynamics}) of two actions $a_1$ and $a_2$ are 0.51 and 0.49 respectively in  state 1, while their responses are 0.99 and 0.01 respectively in state 2. It indicates $a_1$ and $a_2$ have a similar effect at state 1, while $a_1$ is superior than $a_2$ significantly at state 2. However, PNE just save $a_1$ in both states and cannot show the difference. To tackle this problem, we recover the response value for each action by $Q(s, a^g,\tilde A^{-g})$ where $\tilde A$ is PNE. $Q(s, a^g,\tilde A^{-g})$ can be considered as the continuous form extended from PNE and $\tilde a^g=\arg\max\limits_{a^g \in A^g} Q(s, a^g,\tilde A^{-g})$.  Generally speaking, an action with higher $Q(s, a^g,\tilde A^{-g})$ should have a larger probability of being performed. Hence, the objective policy of the performed actions of $g\in G$ can be estimated by:
\begin{align}
\small
\begin{aligned}
 \label{eq:weaklabel}
 \tilde p(a|g,s)=\frac{ Q(s, a,\tilde A^{-g})-\min\limits_{a'\in A^g}{ Q(s, a',\tilde A^{-g})}}{\sum\limits_{a' \in A^g}{Q(s, a',\tilde A^{-g})}-|A^g|  \min\limits_{a'\in A}{ Q(s, a',\tilde A^{-g})}},
\end{aligned}
\end{align}
where $a\in A^g$. Hence, the distribution of  parameterized policy $p^{\theta}$ should be close to $\tilde p(a|g,s)$, and $\theta$ can be learned  by minimum the Kullback-Leibler divergence ($D^{KL}$) between $p^{\theta}$ and $\tilde p$:
\begin{align}
\small
\begin{aligned}
 \label{eq:loss1}
 \theta^*&=\mathop{\arg\min}\limits_{\theta}{D^{KL}(\tilde p || p^{\theta})}\\
        &= \mathop{\arg\max}\limits_{\theta}\sum\limits_{g\in G}{\sum\limits_{a\in A^g}{\log(p_\theta(a|g,s))\times\tilde p(a|g,s)}}.
\end{aligned}
\end{align}
Then, the decentralized policy gradient $\bigtriangledown_{\theta}$ according to $\theta$ can be calculated by (\ref{eq:loss}), and  $\theta$ can be learned by $\theta^{k+1}=\theta^{k}+\eta^{k}\bigtriangledown_{\theta^k}$  where $k$ stands for the $k$th iteration of learning and  $\eta^{k}$ is the learning rate.
\begin{align}
\small
\begin{aligned}
 \label{eq:loss}
\bigtriangledown_{\theta}=\sum\limits_{g\in G}{\sum\limits_{a\in A^g}{\bigtriangledown_{\theta}\log(p_\theta(a|g,s))\times\tilde p(a|g,s)}}.
\end{aligned}
\end{align}

\subsection{Online Update to Q}
As stated in (\ref{eq:loss}) and (\ref{eq:weaklabel}), the parameter $\theta$ can be trained by the policy gradient which is dependent on the $Q(s,A)$. However, $Q(s,A)$ is unknown because the optimal policy is unknown. Fortunately,  although the available demonstration is sub-optimal,  the experts still perform reasonably well. Hence, $Q(s,A)$ can be approximated by $Q^D(s,A)$ at the beginning of learning. $Q^D(s,A)$ is similar to expert trajectory return \cite{Hester2017Deep}, that is, simulating a MDP from $s$ until the terminal state where the reward is available. In this case, the Lemma 1 are introduced.

\noindent\textbf{Lemma 1: }If the PNE (i.e., the output of Alg.~\ref{alg:best-response-dynamics}) are calculated by $Q^D(s,A)$, then $Q^N(s,A)\ge Q^D(s,A)$ where $Q^N(s,A)$ is the state-action value (\ref{eq:ostatevalue}) according to the PNE.

\noindent\textbf{Proof.}
Consider a MDP initialized by $s_0=s$. Since the PNE in Alg.~\ref{alg:best-response-dynamics} are initialized from the demonstration, and the state-action value strictly increases in every iteration, we have:
\begin{align}
\small
\begin{aligned}
 \label{eq:ostatevalue}
 &Q^D(s,A)=\mathbb{E}_{\{s_{t+1}\sim \mathbb{P}_{t+1}, A_0=A,\{A_t\sim p^D\}_{t>0}\}_{t=0}^{T-1}}{\lambda^{T-1} R(s_{T})}\\
         &\leq \mathbb{E}_{\{s_{t+1}\sim \mathbb{P}_{t+1}, A_0=A, A_1\sim\tilde A_1, \{A_t\sim p^D\}_{t>1}\}_{t=0}^{T-1}}{\lambda^{T-1} R(s_{T})}\\
         &\leq \mathbb{E}_{\{s_{t+1}\sim \mathbb{P}_{t+1}, A_0=A, A_1\sim\tilde A_1,A_2\sim\tilde A_2, \{A_t\sim p^D\}_{t>2}\}_{t=0}^{T-1}}{\lambda^{T-1} R(s_{T})}\\
         &...\\
         &\leq \mathbb{E}_{\{s_{t+1}\sim \mathbb{P}_{t+1}, A_0=A, \{A_t\sim \tilde A_t\}_{t>0}\}_{t=0}^{T-1}}{\lambda^{T-1} R(s_{T})= Q^N(s,A)}.
\end{aligned}
\end{align}
%
%
%Then
%
%Since $p^D$ is not optimal, hence $V^D(s)\le V^*(s)$. Note the PNE (i.e., the output of Alg.~\ref{alg:best-response-dynamics}) are the optimal policy based on the $Q$
%
%
%To reduce the gap between $V^D(s)$ and $V^*(s)$, the PNE, i.e.,the output of Alg.~\ref{alg:best-response-dynamics}, are utilized to infer the state-action value instead of the demonstration
%
%
%
%Q as similar to (\ref{eq:optimalreward}):
%\begin{align}
%\small
%\begin{aligned}
% \label{eq:pnereward}
% V^N(s)=\mathbb{E}_{\{s_{t+1}\sim P(s_{t+1}|s_t,\tilde A_t) \}_{t=1}^{T-1}}R(s_{T}|s_{T-1}, A_{T-1}),
%\end{aligned}
%\end{align}
%where $s_1=s$. It can be proved that $V^N(s)\ge V^D(s)$ for every state inductively: Consider a uniform state $s_{T-1}$, due to the PNE in Alg.~\ref{alg:best-response-dynamics} are initialized from the demonstration, and the state reward strictly increases in every iteration. Hence, $R^N(s_{T-1})\ge R^D(s_{T-1})$. This conclusion can be generalized to other states by analogy.

According to Lemma 1, $Q^N(s,a)$ is closer to $Q(s,a)$ than $Q^D(s,a)$ (i.e.,$Q(s,a) \geq Q^N(s,a) \geq Q^D(s,a)$ ). Unfortunately, it is hard to calculate $Q^N(s,a)$ in practice. The calculation of $Q^N(s,a)$ relies  Alg.~\ref{alg:best-response-dynamics} in each state transition of (\ref{eq:ostatevalue}). It would be time consuming and leads to the learning inefficient. To reduce computational cost, $Q^{\theta}(s,A)$ is used to replace $Q^N(s,A)$:
\begin{align}
\small
\begin{aligned}
 \label{eq:netreward}
 Q^{\theta}(s,A)=\mathbb{E}_{\{s_{t+1}\sim \mathbb{P}_{t+1}, A_t=\arg\max\limits_{a}p^\theta\}_{t=0}^{T-1}}{\lambda^{T-1} R(s_{T})},
\end{aligned}
\end{align}
 where $s_0=s$ and $A_0=A$.
In the learning procedure, the DPN is initialized randomly and can't model the PNE effectively at some cases during learning procedure. To ensure that the $Q$ robust, it is calculated as (\ref{eq:finalreward}) in the proposed method:
 \begin{align}
\small
\begin{aligned}
 \label{eq:finalreward}
 Q(s,A) \approx \max( Q^{\theta}(s,A), Q^{D}(s,A)).
\end{aligned}
\end{align}

Alg.~\ref{alg:learning} concludes the proposed learning algorithm. It is based on self-improvement. That is, the $Q(s,A)$ are used to guide to update the network and the network can improve $Q(s,A)$ in turn. Note that the joint action-state spaces of multiple agents are extremely large and the feasible action-states only occupy a very small portion. To ensure that the train samples effective and abundant, a staged exploration method is utilized. At the early stage of learning, the exploration actions are performed according to the PNE to collect effective train samples efficiently. Then the actions are guided by the network to cover more cases.
\begin{algorithm}[t]
\small
	\KwIn{The demonstration policy $p^D(a|g,s)$ for each agent $g\in G$ and state $s$.}
    Initialize $\theta$ randomly\\
    \While{Non-convergence}
    {
        Initialize $s_0$ randomly.\\
        \For{$t=0,...,T$}
        {
           Define the $Q$ function by (\ref{eq:finalreward}).\\
            Calculate the PNE $\tilde A$ by Alg.~\ref{alg:best-response-dynamics}.\\
             \For{$g \in G$}
                {
                    Extract the feature vector.\\
                    Calculate the objective policy by (\ref{eq:weaklabel}). \\
                }
             \If{At the early stage of learning}
             {
                Explore the environment by the PNE: $s_{t+1}\sim \mathbb{P}(s_{t+1}|s_t,  \tilde A_t)$;
             }
             \Else
             {
                Explore the environment by the DPN: $s_{t+1}\sim \mathbb{P}(s_{t+1}|s_t, \{a^g \sim  p^{\theta}(a|f(g,s_t)) \})$;
             }
             Update $\theta$ by (\ref{eq:loss}).
        }
    }
\caption{The proposed RL algorithm.}
\label{alg:learning}
\end{algorithm}
\vspace{-3mm}
\section{Experiments}

\label{sec_experiment}
\subsection{Combat in RTS Game }
The proposed method are tested in combat RTS game where the task is to coordinate multiple allied agents to defeat their enemies controlled by an opponent in a real-time scenario. The combat game has a high requirement for multi-agent cooperation and has been widely utilized to test multi-agent based methods \cite{usunier2016episodic,peng2017multiagent,Xiangyu2017Revisiting,NIPS2017_7217,foerster2017counterfactual,pmlr-v80-rashid18a,churchill2012fast,churchill2013portfolio,Lelis2017Stratified}.
From a machine learning point of view, combat game provides a challenging environment to test AI algorithms because its state-action space is extremely large and the time allowed for planning is on the order of milliseconds. For example, in a specific 15 vs. 15 scenario, the number of points in joint state-action space is almost $10^{500}$ which is much larger than for Go ($10^{170}$) \cite{silver2016mastering}. The proposed method is testing using SparCraft \cite{churchill2012fast}, which is a simulator of the StarCraft local combat game and is widely adopted to test combat game algorithms \cite{churchill2012fast,churchill2013portfolio,Lelis2017Stratified}.
It is chosen as the experimental platform because of its effective forward simulation function which can make the RL algorithm, especially Alg.~\ref{alg:best-response-dynamics}, more efficiently. In addition, the SparCraft implements several different rule-based heuristics which can be used as our sub-optimal demonstrations.
The source code and the trained models will be opened to facilitate further studies of this problem.
In our experiment, the combat game is modeled as a sparse-reward task:  the reward is only defined at the terminal state (\ref{eq:actionreward}) where all agents of one player have been wiped out. In (\ref{eq:actionreward}), $ G^0$ are allied agents and $ G^1$ are enemies.
\begin{align}
\small
\begin{aligned}
 \label{eq:actionreward}
 R(s_{T})=\sum\limits_{g\in G^0}{hp(g)}-\sum\limits_{g\in G^1}{hp(g)},
\end{aligned}
\end{align}

%In SparCraft, the unit properties such as weapons damage are same to StarCraft, while the fog of war, collisions and unit acceleration are not implemented.
\subsection{Settings}
The decay factor $\lambda$ can be set to any positive number because it doesn't affect the objective policy (\ref{eq:weaklabel}). The iterations in Alg.~\ref{alg:best-response-dynamics} is set to 10 in our experiments. The win rate over 100 battles and the normalized rewards at terminal states \footnote{$ R(s_{T})$/(The sum of all allied agents' $hp$ at the initial state)} are used as evaluation metrics.
%The learning rate is equal to 0.01 for both learning stages. The batch size is 128 in the first stage and the stage 2 learning is terminated after about 200 battles.
We utilize 2 rule-based heuristics implemented in SparCraft as our demonstrators, including: (1) \textbf{Attack-Closest} (\emph{c}) where agents attack the closest enemy within weapon range, and any agent not within the range of any enemy moves toward the closest enemy, and (2) \textbf{Attack-Weakest} (\emph{w}):where agents attack the enemy  with minimum remaining $hp$ within weapon range, and an agent not within the range of any enemy moves toward the closest enemy.
%These two demonstrators are both obviously sub-optimal and need to be improved.
 In our experiments, two types of demonstrations are given for each demonstrator: the heuristic simulator and the observed data. The data are used to recover the policy by imitation learning (\ref{eq:imtationloss}) and the learned policy is further used to calculate $Q^D$ (\ref{eq:finalreward}).
 As same as most MARL works \cite{usunier2016episodic,peng2017multiagent,Xiangyu2017Revisiting,foerster2017counterfactual}, the opponent's policy can be simulated as a part of environment in learning. It is reasonable because our goal is to learn multi-agent cooperation rather than competition.  In all the tested combat scenarios, two base points in the combat area are chosen for allied and enemy units respectively. The positions of each player's agents are initialized randomly within a range of the appropriate base point. That is, the scenarios are different in each combat. Our experiments are conducted on a cross-validation setting that the model learned from  demonstrator \emph{c} will fight with \emph{w} and vice versa.
The Terrain Marine (m) is chosen as the test agent type. The test combat scenarios differ in different scales and difficulties: a small scale combat m5v5, a large scale combat m30v30, and two unbalanced combats m18v20 and m24v30 where we control 18 (24) Marines against 20 (30) Marines and our force is much weaker than enemy. These combats, especially the unbalanced combats, require the multiple agents to cooperate effectively to defeat the enemy. The models are trained on GeForce GTX 1080 and tested
on a desktop PC with one 2.4 GHz CPU and 8G RAM running in C++. The network can make decision in real time because it only needs forwarding a simple network only once in making each decision.
\vspace{-1mm}
\subsection{Feature and Network Architecture}

In the experiments, the action space $A$ contains two types of discrete actions: move[$directions$] and attack[$enemy\_ids$], where $directions$ is set to 4 corresponding to left, right, up and down, and $enemy\_ids$ is the number of enemy units in the start game state of the train combat scenario.
To an allied agent $g$, the list [$enemy\_ids$] is the list of attack targets of $g$, and it is arranged in two parts sequentially: the enemy agents inside and outside the weapon range of $g$ respectively. In each part, the list is arranged by the ascending order of the remaining hit points $hp$.
The feature vector ($f(g,s)$) is concatenated of 4 parts: (1) The properties of $g$ such as maximum $hp$, velocity, weapon damage,  maximum cooldown ($cd$, the number of frames to wait before being able to attack again), damage per frame, current $hp$, position (coordinates on the map) and current $cd$. (2) The properties of the enemy agents which are concatenated by the order of [$enemy\_ids$]. (3) The properties of the allied agents which are concatenated by ascending order of the distance from the allied agent to the corresponding agent $g$.  (4) The statistical properties of the allied and enemy units respectively including the mean, the minimum and the maximum value of the current hit points and the center positions. The positions of the agents except  $g$ are all the relative coordinates respect to $g$.
If the number of allied and enemy agents in a game state is smaller than initial game state, that is, some agents have been wiped out, then these eliminated agents  are replaced by nominal agents whose properties are all 0 to make the length of the feature vector identical. In test combat scenarios, the legal action with maximum probability according to the DPN is chosen to be performed. The network is composed of 4 fully connected (FC) layers, 3 batch normalization layers following the first 3 FC layers respectively and a softmax layer to output the probability distribution over the whole action space $A$. The widths of the FC layers are 256, 128, 128 and $directions + enemy\_ids$ respectively. The leaky rectified linear function \cite{maas2013rectifier} is used as an activation function for the first 3 FC layers.

\subsection{Comparison Results}
Under this setting that the demonstrations are given, the compared methods can be categorised into two groups: (1) Heuristic-based search methods, including UCT \cite{churchill2013portfolio}, Alpha-Beta (A-B) search  \cite{churchill2013portfolio} and profile search \cite{churchill2013portfolio}. The former two methods use the demonstration policy to guide game tree search. Profile search is proposed to search the optimal heuristic for each agent and two demonstration policy are both used as the candidate policies in the experiments. (2) Deep LfD methods, including DQfQ \cite{Hester2017Deep}, DDPGfD \cite{Ve2017Leveraging} and OGTL \cite{OGTL}. These methods are re-implemented with the same feature and experimental setting for fair comparison. The first 2 methods are all designed for single agent, while they can be easily extended to multi-agent setting by combining with IQL \cite{Tan1993Multi}. OGTL is the latest LfD method designed for multi-agent learning to our knowledge.  It utilize the human-made data to pre-train the network and then fine-tuned by RL where rewards are available at every state. For fair comparison, OGTL is pre-trained by the observed data and $Q$ is calculated as (\ref{eq:finalreward}).

The comparative results presented in Table 1 are the basis of the following:
(1) The search based methods UCT and Alpha-Beta can improve demonstration policy clearly in balanced combats, while the improvement is very limited in unbalanced combats. It is difficult to search the effective policy from the extremely large game tree under real-time limitation (40ms). These methods  can only search around the demonstration policy. Profile search select the heuristic without additional exploration, hence it can't find effective cooperative policy because the used heuristics are both sub-optimal.  (2) DQfQ  suffer the poor performance because its learning policy  is constrained to be close to the demonstration, while it limits the model when the demonstration is sub-optimal. DDPGfD and OGTL perform  well in balanced combats, and suffer poorly in unbalanced combats because the cooperation is ignored during learning.
(3) The proposed method outperforms other methods as well as the demonstration policy significantly in all testing scenarios. The advantage is more obvious in unbalanced combats. For example, in m24v30 with demonstration \emph{w}, the win rates of other method are all bellow 0.10, while the proposed method can get 0.58. These advantages indicates the proposed method can significantly improve the demonstration and learn multi-agent cooperation effectively.

\begin{table}[ht]
\centering
\fontsize{7pt}{8pt}\selectfont
\scriptsize
\begin{tabular}{|c|c c |c c| c c| c c|}
            %{|p{1.5cm}|p{1.5cm}<{\centering} p{1.5cm}<{\centering} p{1.5cm}<{\centering}|p{1.5cm}<{\centering}|}
			 \hline
        \centering \scriptsize{Scenarios} & \multicolumn{2}{|c|}  {\scriptsize{m5v5}} & \multicolumn{2}{|c|} {\scriptsize{m30v30}} & \multicolumn{2}{|c|} {\scriptsize{m18v20}} & \multicolumn{2}{|c|} {\scriptsize{m24v30}}\\
        \hline
        \centering Metrics & W & R & W &R & W&R& W&R\\
        \hline
            \centering D\_\emph{c}& 0.43 & -0.02 & 0.75 &0.18 &0.32&-0.11& 0.12&-0.26\\
            \centering UCT\_\emph{c}& 0.93 & 0.06 & 0.88 &0.13 &0.40&-0.14&0.13 &-0.25\\
            \centering A-B\_\emph{c}& 0.96 & 0.07 & 0.85 &0.16 &0.36&-0.17&0.12 &-0.24\\
            \centering Profile & 0.84 & 0.14 & 0.99 &0.33 &0.61&0.06&0.30 &-0.11\\
            \hline
            \centering DQfQ\_\emph{c}& 0.54 & 0.01 & 0.81 &0.23 &0.37&-0.10& 0.14 &-0.20\\
            \centering DDPGfD\_\emph{c} & 0.86 & 0.21 & 0.95 &0.05 &0.48&-0.19&0.16 &-0.17\\
            \centering OGTL\_\emph{c}& 0.89 & 0.24 & 0.93 &0.16 &0.38&-0.05& 0.13 &-0.21\\
            \hline
            \centering Ours(O)\_\emph{c} & 0.94 & 0.43 & 0.97 &0.49 &0.86&0.22& 0.73&0.15\\
            \centering Ours(H)\_\emph{c} &\textbf{ 0.98} & \textbf{0.46} & \textbf{1.00} &\textbf{0.51} &\textbf{0.87}&\textbf{0.24}&\textbf{0.76} &\textbf{0.21}\\
            \hline
            \hline
            \centering D\_\emph{w}& 0.61 & 0.02 & 0.13 &-0.23 &0.03&-0.36&0.00 &-0.49\\
            \centering UCT\_\emph{w}& 0.93 & 0.01 & 0.73 &0.09 &0.37&-0.19&0.00 &-0.36\\
            \centering A-B\_\emph{w}& 0.92 & 0.02 & 0.71 &0.09 &0.34&-0.26&0.00 &-0.38\\
            \centering Profile & 0.91 & 0.16 & 0.96 &0.21 &0.39&-0.08&0.02 &-0.29\\
            \hline
            \centering DQfQ\_\emph{c}& 0.69 & 0.08 & 0.34 &-0.13 &0.27&-0.19& 0.04 &-0.35\\
            \centering DDPGfD\_\emph{w}& 0.78 & 0.15 & 0.75 &0.09 &0.69&0.13& 0.08 &-0.31\\
            \centering OGTL\_\emph{w}& 0.81 & 0.14 & 0.73 &0.15 &0.76&0.19& 0.06 &-0.38\\
            \hline
            \centering Ours(O)\_\emph{w} & 0.96 & 0.89 & 0.81 &0.72 &0.68&0.15&0.52&0.01\\
            \centering Ours(H)\_\emph{w} & \textbf{0.99} & \textbf{0.48} & \textbf{0.84} &\textbf{0.26} &\textbf{0.71}&\textbf{0.16}&\textbf{0.58} &\textbf{0.03}\\
            \hline
        \end{tabular}
\caption{A comparison with other demonstration based approaches where ``\_\emph{c}''(``\_\emph{w}'') means the used demonstration policy is \emph{c} (\emph{w}). (H) and (O) stand for the heuristic simulator and observed data respectively. ``D'' is the demonstration policy. The best result for the given scenario is in bold.}
\vspace{-4mm}
\label{tab:rl}
\end{table}

\subsection{Further Evaluation}
\vspace{-1mm}
We perform ablation experiments to validate two key elements of the proposed method. First, the proposed method (\ref{eq:loss}) is compared with two other MARL methods IQL \cite{Tan1993Multi} and COMA \cite{foerster2017counterfactual}. Since they are both designed for the case in which the reward can be calculated at every state, the $Q$ used in these methods are all calculated from (\ref{eq:finalreward}) for fair comparison. As shown in Table \ref{tab:r2}, IQL performs well in balanced scenarios while fairs poorly in unbalanced scenarios. The reason is that IQL learn each agent independently without effective cooperation. The win rate of COMA is fairly high in small balanced battle but apparently fell in large and unbalanced battles. Different with COMA, the proposed method learn decentralized policies according to Nash Equilibrium. The guidance of Nash Equilibrium reduce the multi-agent exploration space and leads to more effective and stable cooperation learning.

Then, the contribution of ``online update of $Q$'' is evaluated. In this experiment, the proposed method is compared with two ablation methods:  ``Ours$^D$'' and ``Ours$^{\theta}$'' where $Q$ are replaced by $Q^D$ and $Q^{\theta}$ respectively throughout the learning procedure. From the results in Table~\ref{tab:r3}, it is evident that: (1) ``Ours$^D$'' performs well, thus demonstrating the effectiveness of using sub-optimal demonstration. (2) The performance of ``Ours$^\theta$'' is very poor because the network cannot provide an effective $Q$ when the rewards are sparse.  (3) ``Ours'' outperforms ``Ours$^D$'' and ``Ours$^\theta$'' in all cases which demonstrates (\ref{eq:finalreward}) contribute positively.
\begin{table}[ht]
\centering
\fontsize{7pt}{8pt}\selectfont
\scriptsize
\begin{tabular}{|c|c c| c c| c c |c c|}
            %{|p{1.5cm}|p{1.5cm}<{\centering} p{1.5cm}<{\centering} p{1.5cm}<{\centering}|p{1.5cm}<{\centering}|}
		 \hline
        \centering \scriptsize{Scenarios} & \multicolumn{2}{|c|}  {\scriptsize{m5v5}} & \multicolumn{2}{|c|} {\scriptsize{m30v30}} & \multicolumn{2}{|c|} {\scriptsize{m18v20}} & \multicolumn{2}{|c|} {\scriptsize{m24v30}}\\
        \hline
        \centering Metrics & W & R & W &R & W&R& W&R\\
        \hline
			
            \centering IQL\_\emph{c}& 0.89 & 0.28 & 0.88 &0.33 &0.69& 0.08 & 0.47&-0.05\\
            \centering COMA\_\emph{c}& \textbf{0.99} & 0.41 & 0.77 &0.18 &0.61&0.17& 0.26& -0.11\\
            \hline
             \centering Ours\_\emph{c} & 0.98 & \textbf{0.46} & \textbf{1.00} &\textbf{0.51} &\textbf{0.87}&\textbf{0.24}&\textbf{0.76} &\textbf{0.21}\\
            \hline
        \end{tabular}
\caption{A comparison with different multi-agent reinforcement learning methods where $Q$ is calculated by (\ref{eq:finalreward}) using the heuristic simulator.}
\label{tab:r2}
\end{table}

\begin{table}[ht]
\centering
\fontsize{7pt}{8pt}\selectfont
\scriptsize
\begin{tabular}{|c|c c| c c| c c |c c|}
            %{|p{1.5cm}|p{1.5cm}<{\centering} p{1.5cm}<{\centering} p{1.5cm}<{\centering}|p{1.5cm}<{\centering}|}
		 \hline
        \centering \scriptsize{Scenarios} & \multicolumn{2}{|c|}  {\scriptsize{m5v5}} & \multicolumn{2}{|c|} {\scriptsize{m30v30}} & \multicolumn{2}{|c|} {\scriptsize{m18v20}} & \multicolumn{2}{|c|} {\scriptsize{m24v30}}\\
        \hline
        \centering Metrics & W & R & W &R & W&R& W&R\\
        \hline
            \centering Ours$^{\theta}$\_\emph{c}&0.02 & -0.51 & 0.01 & -0.63 & 0.00&-0.67&0.00 &-0.72\\
            \centering Ours$^D$\_\emph{c}& 0.96 & 0.38 & 0.96 &0.27 &0.79&0.18&0.67 & 0.16\\
            \hline
            \centering Ours\_\emph{c} & \textbf{0.98} & \textbf{0.46} & \textbf{1.00} &\textbf{0.51} &\textbf{0.87}&\textbf{0.24}&\textbf{0.76} &\textbf{0.21}\\
            \hline
        \end{tabular}
\caption{A comparison with different calculations of $Q$. (H) is used as demonstration.}
\vspace{-3mm}
\label{tab:r3}
\end{table}

\begin{figure}[htp]
		\centering
		\setlength{\belowcaptionskip}{0pt}
		\includegraphics[width = 0.6\linewidth]{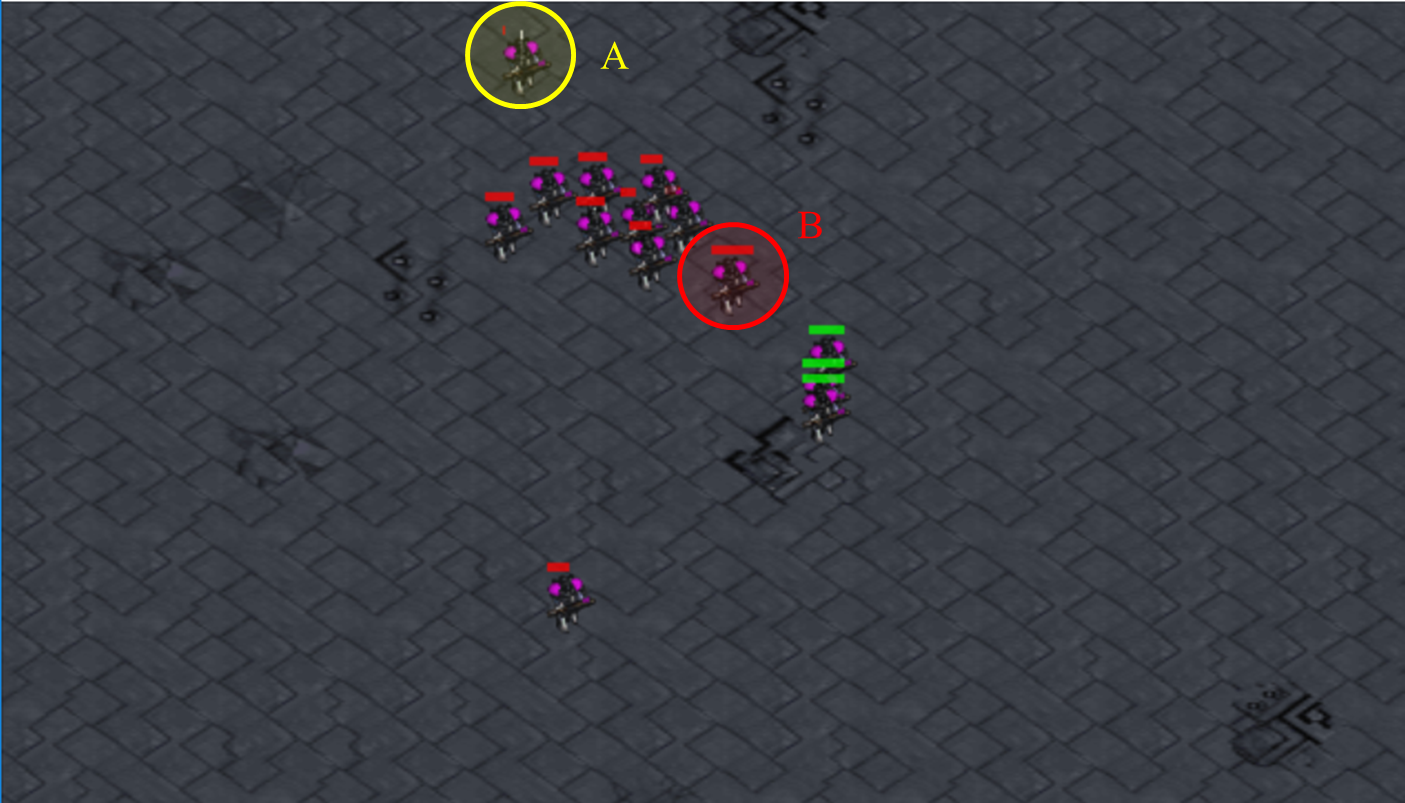}
		\caption{ A typical battle scene where the units with red hp bar are controlled by the learned policy and the green hp bar units are opponent units.}
        \label{fig:combatexample}
\end{figure}

To better understand the proposed method, here we give two intuitive examples to show what the proposed method have learned\footnote{More demos will be released with the source codes.}: (1) Cover and withdraw. Fig~\ref{fig:combatexample} is a typical battle scene from m18v20. We can find that the learned policy will keep most units engaged in combat to maximize the damage. Also, the healthy unit (B) will step forward to cover other ally units and the near death unit (A) will escape from the battlefield to keep alive. (2) Fire allocation to maximum team damage. The used demonstration policies focus fire to eliminate a enemy agent quickly, which may lead ``over kill''. For example, coordinate all agents to attack a near death unit. In this case, most attacks will be invalid because the target will be eliminated by one attack. The ratio of the invalid attacks are counted to estimate this situation. In m30v30 scenario, the ratio of demonstration $w$ is 43\%, while it is only 6\% to the learned policy. It shows the proposed method can allocate fire more effectively.
\vspace{-3mm}
\section{Conclusion}
\label{sec_conclusion}
This paper introduces a method to learn multi-agent cooperation from sub-optimal demonstration.  A self-improving method is proposed to learn a decentralized policy: On the one hand, the state-action value are initialized by the demonstration and updated by the learned strategy to improve its sub-optimality. On the other hand,  the Nash Equilibrium strategies are found by the current state-action value to learn better policy. The proposed method is evaluated on the combat of RTS game which has a high requirement for multi-agent cooperation.  Extensive experimental results on various combat scenarios demonstrate that the proposed method can learn the multi-agent cooperation effectively and it significantly outperforms many state-of-the-art demonstration based approaches.
{%\small
\small
\linespread{0.5}\selectfont
\bibliographystyle{aaai}
\bibliography{fixbib}

\begin{thebibliography}{}

\bibitem[\protect\citeauthoryear{Busoniu, Babuska, and
  De~Schutter}{2008}]{Busoniu2008A}
Busoniu, L.; Babuska, R.; and De~Schutter, B.
\newblock 2008.
\newblock A comprehensive survey of multiagent reinforcement learning.
\newblock {\em IEEE Transactions on Systems Man and Cybernetics}
  38(2):156--172.

\bibitem[\protect\citeauthoryear{Cao \bgroup et al\mbox.\egroup
  }{2012}]{Cao2012An}
Cao, Y.; Yu, W.; Ren, W.; and Chen, G.
\newblock 2012.
\newblock An overview of recent progress in the study of distributed
  multi-agent coordination.
\newblock {\em IEEE Transactions on Industrial Informatics} 9(1):427--438.

\bibitem[\protect\citeauthoryear{Chen and etc.}{2016}]{chen2017decentralized}
Chen, Y.~F., and etc.
\newblock 2016.
\newblock Decentralized non-communicating multiagent collision avoidance with
  deep reinforcement learning.
\newblock In {\em Proc. ICRA}.

\bibitem[\protect\citeauthoryear{Churchill and
  Buro}{2013}]{churchill2013portfolio}
Churchill, D., and Buro, M.
\newblock 2013.
\newblock Portfolio greedy search and simulation for large-scale combat in
  starcraft.
\newblock In {\em Proc. CIG},  1--8.
\newblock IEEE.

\bibitem[\protect\citeauthoryear{Churchill, Saffidine, and
  Buro}{2012}]{churchill2012fast}
Churchill, D.; Saffidine, A.; and Buro, M.
\newblock 2012.
\newblock Fast heuristic search for rts game combat scenarios.
\newblock In {\em Proc. AIIDE},  112--117.

\bibitem[\protect\citeauthoryear{Everett, Chen, and
  How}{2018}]{Everett2018Motion}
Everett, M.; Chen, Y.~F.; and How, J.~P.
\newblock 2018.
\newblock Motion planning among dynamic, decision-making agents with deep
  reinforcement learning.
\newblock {\em arXiv}.

\bibitem[\protect\citeauthoryear{Foerster \bgroup et al\mbox.\egroup
  }{2017}]{pmlr-v70-foerster17b}
Foerster, J.; Nardelli, N.; Farquhar, G.; Afouras, T.; Torr, P. H.~S.; Kohli,
  P.; and Whiteson, S.
\newblock 2017.
\newblock Stabilising experience replay for deep multi-agent reinforcement
  learning.
\newblock In {\em Proc. ICML}.

\bibitem[\protect\citeauthoryear{Foerster \bgroup et al\mbox.\egroup
  }{2018}]{foerster2017counterfactual}
Foerster, J.; Farquhar, G.; Afouras, T.; Nardelli, N.; and Whiteson, S.
\newblock 2018.
\newblock Counterfactual multi-agent policy gradients.
\newblock In {\em Proc. AAAI}.

\bibitem[\protect\citeauthoryear{Guestrin, Koller, and
  Parr}{2002}]{Guestrin2001Multiagent}
Guestrin, C.; Koller, D.; and Parr, R.
\newblock 2002.
\newblock Multiagent planning with factored mdps.
\newblock In {\em Proc. NIPS}.

\bibitem[\protect\citeauthoryear{han Chang, Ho, and
  Kaelbling}{2004}]{NIPS2003_2476}
han Chang, Y.; Ho, T.; and Kaelbling, L.~P.
\newblock 2004.
\newblock All learning is local: Multi-agent learning in global reward games.
\newblock In {\em Proc. NIPS}.

\bibitem[\protect\citeauthoryear{Hester and etc.}{2017}]{Hester2017Deep}
Hester, T.~a., and etc.
\newblock 2017.
\newblock Deep q-learning from demonstrations.
\newblock In {\em Proc. AAAI}.

\bibitem[\protect\citeauthoryear{Ho and Ermon}{2016}]{Ho2016Generative}
Ho, J., and Ermon, S.
\newblock 2016.
\newblock Generative adversarial imitation learning.
\newblock In {\em Proc. NIPS}.

\bibitem[\protect\citeauthoryear{Hu \bgroup et al\mbox.\egroup }{2018}]{OGTL}
Hu, Y.; Li, J.; Li, X.; Pan, G.; and Xu, M.
\newblock 2018.
\newblock Knowledge-guided agent-tactic-aware learning for starcraft
  micromanagement.
\newblock In {\em Proc. IJCAI}.

\bibitem[\protect\citeauthoryear{Kang, Jie, and Feng}{2018}]{pmlr-v80-kang18a}
Kang, B.; Jie, Z.; and Feng, J.
\newblock 2018.
\newblock Policy optimization with demonstrations.
\newblock In {\em Proc. ICML}.

\bibitem[\protect\citeauthoryear{Kok and Vlassis}{2006}]{Kok2006Collaborative}
Kok, J.~R., and Vlassis, N.
\newblock 2006.
\newblock Collaborative multiagent reinforcement learning by payoff
  propagation.
\newblock {\em Journal of Machine Learning Research} 7(1):1789--1828.

\bibitem[\protect\citeauthoryear{Kong \bgroup et al\mbox.\egroup
  }{2017}]{Xiangyu2017Revisiting}
Kong, X.; Xin, B.; Liu, F.; and Wang, Y.
\newblock 2017.
\newblock Revisiting the master-slave architecture in multi-agent deep
  reinforcement learning.
\newblock {\em arXiv preprint arXiv:1712.07305}.

\bibitem[\protect\citeauthoryear{Lanctot \bgroup et al\mbox.\egroup
  }{2017}]{NIPS2017_7007}
Lanctot, M.; Zambaldi, V.; Gruslys, A.; Lazaridou, A.; Tuyls, k.; Perolat, J.;
  Silver, D.; and Graepel, T.
\newblock 2017.
\newblock A unified game-theoretic approach to multiagent reinforcement
  learning.
\newblock In {\em Proc. NIPS}.

\bibitem[\protect\citeauthoryear{LeCun, Bengio, and
  Hinton}{2015}]{lecun2015deep}
LeCun, Y.; Bengio, Y.; and Hinton, G.
\newblock 2015.
\newblock Deep learning.
\newblock {\em Nature} 521(7553):436--444.

\bibitem[\protect\citeauthoryear{Lelis}{2017}]{Lelis2017Stratified}
Lelis, L. H.~S.
\newblock 2017.
\newblock Stratified strategy selection for unit control in real-time strategy
  games.
\newblock In {\em Proc. IJCAI},  3735--3741.

\bibitem[\protect\citeauthoryear{Levine \bgroup et al\mbox.\egroup
  }{2016}]{Levine2016End}
Levine, S.; Finn, C.; Darrell, T.; and Abbeel, P.
\newblock 2016.
\newblock End-to-end training of deep visuomotor policies.
\newblock {\em Journal of Machine Learning Research} 17(1):1334--1373.

\bibitem[\protect\citeauthoryear{Lin, Beling, and
  Cogill}{2018}]{Lin2018Multiagent}
Lin, X.; Beling, P.~A.; and Cogill, R.
\newblock 2018.
\newblock Multiagent inverse reinforcement learning for two-person zero-sum
  games.
\newblock {\em IEEE Transactions on Games} 10(1):56--68.

\bibitem[\protect\citeauthoryear{Lowe \bgroup et al\mbox.\egroup
  }{2017}]{NIPS2017_7217}
Lowe, R.; WU, Y.; Tamar, A.; Harb, J.; Pieter~Abbeel, O.; and Mordatch, I.
\newblock 2017.
\newblock Multi-agent actor-critic for mixed cooperative-competitive
  environments.
\newblock In {\em Proc. NIPS}.

\bibitem[\protect\citeauthoryear{Maas, Hannun, and
  Ng}{2013}]{maas2013rectifier}
Maas, A.~L.; Hannun, A.~Y.; and Ng, A.~Y.
\newblock 2013.
\newblock Rectifier nonlinearities improve neural network acoustic models.
\newblock In {\em Proc. ICML}, volume~30, ~3.

\bibitem[\protect\citeauthoryear{Mnih \bgroup et al\mbox.\egroup
  }{2015}]{mnih2015human}
Mnih, V.; Kavukcuoglu, K.; Silver, D.; Rusu, A.~A.; Veness, J.; Bellemare,
  M.~G.; Graves, A.; Riedmiller, M.; Fidjeland, A.~K.; Ostrovski, G.; et~al.
\newblock 2015.
\newblock Human-level control through deep reinforcement learning.
\newblock {\em Nature} 518(7540):529.

\bibitem[\protect\citeauthoryear{Ng and Russell}{2000}]{Ng2000Algorithms}
Ng, A.~Y., and Russell, S.~J.
\newblock 2000.
\newblock Algorithms for inverse reinforcement learning.
\newblock In {\em Proc. ICML}.

\bibitem[\protect\citeauthoryear{Omidshafiei \bgroup et al\mbox.\egroup
  }{2017}]{pmlr-v70-omidshafiei17a}
Omidshafiei, S.; Pazis, J.; Amato, C.; How, J.~P.; and Vian, J.
\newblock 2017.
\newblock Deep decentralized multi-task multi-agent reinforcement learning
  under partial observability.
\newblock In {\em Proc. ICML}.

\bibitem[\protect\citeauthoryear{Peng \bgroup et al\mbox.\egroup
  }{2017}]{peng2017multiagent}
Peng, P.; Yuan, Q.; Wen, Y.; Yang, Y.; Tang, Z.; Long, H.; and Wang, J.
\newblock 2017.
\newblock Multiagent bidirectionally-coordinated nets for learning to play
  starcraft combat games.
\newblock {\em arXiv preprint arXiv:1703.10069}.

\bibitem[\protect\citeauthoryear{Rashid \bgroup et al\mbox.\egroup
  }{2018}]{pmlr-v80-rashid18a}
Rashid, T.; Samvelyan, M.; Schroeder, C.; Farquhar, G.; Foerster, J.; and
  Whiteson, S.
\newblock 2018.
\newblock {QMIX}: Monotonic value function factorisation for deep multi-agent
  reinforcement learning.
\newblock In {\em Proc. ICML}.

\bibitem[\protect\citeauthoryear{Reddy \bgroup et al\mbox.\egroup
  }{2012}]{Reddy2012Inverse}
Reddy, T.~S.; Gopikrishna, V.; Zaruba, G.; and Huber, M.
\newblock 2012.
\newblock Inverse reinforcement learning for decentralized non-cooperative
  multiagent systems.
\newblock In {\em Systems, Man, and Cybernetics}.

\bibitem[\protect\citeauthoryear{Ross, Gordon, and Bagnell}{2010}]{Ross2010A}
Ross, S.; Gordon, G.~J.; and Bagnell, J.~A.
\newblock 2010.
\newblock A reduction of imitation learning and structured prediction to
  no-regret online learning.
\newblock {\em Aistats} abs/1011.0686.

\bibitem[\protect\citeauthoryear{Schaal}{1997}]{schaal1997learning}
Schaal, S.
\newblock 1997.
\newblock Learning from demonstration.
\newblock In {\em Proc.NIPS}.

\bibitem[\protect\citeauthoryear{Silver \bgroup et al\mbox.\egroup
  }{2016}]{silver2016mastering}
Silver, D.; Huang, A.; Maddison, C.~J.; Guez, A.; Sifre, L.; Van Den~Driessche,
  G.; Schrittwieser, J.; Antonoglou, I.; Panneershelvam, V.; Lanctot, M.;
  et~al.
\newblock 2016.
\newblock Mastering the game of go with deep neural networks and tree search.
\newblock {\em Nature} 529(7587):484--489.

\bibitem[\protect\citeauthoryear{Silver \bgroup et al\mbox.\egroup
  }{2017}]{Silver2017Mastering}
Silver, D.; Schrittwieser, J.; Simonyan, K.; Antonoglou, I.; Huang, A.; Guez,
  A.; Hubert, T.; Baker, L.; Lai, M.; and Bolton, A.
\newblock 2017.
\newblock Mastering the game of go without human knowledge.
\newblock {\em Nature} 550(7676):354--359.

\bibitem[\protect\citeauthoryear{Stone and Veloso}{2000}]{Stone2000Multiagent}
Stone, P., and Veloso, M.
\newblock 2000.
\newblock Multiagent systems: A survey from a machine learning perspective.
\newblock  345--383.

\bibitem[\protect\citeauthoryear{Sukhbaatar and
  Fergus}{2016}]{sukhbaatar2016learning}
Sukhbaatar, S., and Fergus, R.
\newblock 2016.
\newblock Learning multiagent communication with backpropagation.
\newblock In {\em Proc. NIPS},  2244--2252.

\bibitem[\protect\citeauthoryear{Sun \bgroup et al\mbox.\egroup
  }{2017}]{Sun2017Deeply}
Sun, W.; Venkatraman, A.; Gordon, G.~J.; Boots, B.; and Bagnell, J.~A.
\newblock 2017.
\newblock Deeply {A}ggre{V}a{T}e{D}: Differentiable imitation learning for
  sequential prediction.
\newblock In {\em Proc. ICML}.

\bibitem[\protect\citeauthoryear{Syed and Schapire}{2008}]{Syed2008A}
Syed, U., and Schapire, R.~E.
\newblock 2008.
\newblock A game-theoretic approach to apprenticeship learning.
\newblock In {\em Proc. NIPS}.

\bibitem[\protect\citeauthoryear{Syed, Bowling, and
  Schapire}{2008}]{Syed2008Apprenticeship}
Syed, U.; Bowling, M.; and Schapire, R.~E.
\newblock 2008.
\newblock Apprenticeship learning using linear programming.
\newblock In {\em Proc. ICML}.

\bibitem[\protect\citeauthoryear{Tampuu \bgroup et al\mbox.\egroup
  }{2017}]{tampuu2017multiagent}
Tampuu, A.; Matiisen, T.; Kodelja, D.; Kuzovkin, I.; Korjus, K.; Aru, J.; Aru,
  J.; and Vicente, R.
\newblock 2017.
\newblock Multiagent cooperation and competition with deep reinforcement
  learning.
\newblock {\em PloS one} 12(4).

\bibitem[\protect\citeauthoryear{Tan}{1993}]{Tan1993Multi}
Tan, M.
\newblock 1993.
\newblock Multi-agent reinforcement learning: Independent vs. cooperative
  agents.
\newblock {\em Machine Learning Proceedings}  330--337.

\bibitem[\protect\citeauthoryear{Tuyls and Weiss}{2012}]{Tuyls2012Multiagent}
Tuyls, K., and Weiss, G.
\newblock 2012.
\newblock Multiagent learning: Basics, challenges, and prospects.
\newblock {\em Ai Magazine} 33(3):41--52.

\bibitem[\protect\citeauthoryear{Usunier \bgroup et al\mbox.\egroup
  }{2016}]{usunier2016episodic}
Usunier, N.; Synnaeve, G.; Lin, Z.; and Chintala, S.
\newblock 2016.
\newblock Episodic exploration for deep deterministic policies: An application
  to starcraft micromanagement tasks.
\newblock {\em arXiv preprint arXiv:1609.02993}.

\bibitem[\protect\citeauthoryear{Vecerik \bgroup et al\mbox.\egroup
  }{2017}]{Ve2017Leveraging}
Vecerik, M.; Hester, T.; Scholz, J.; Wang, F.; Pietquin, O.; Piot, B.; Heess,
  N.; Roth?rl, T.; Lampe, T.; and Riedmiller, M.
\newblock 2017.
\newblock Leveraging demonstrations for deep reinforcement learning on robotics
  problems with sparse rewards.
\newblock {\em arXiv}.

\bibitem[\protect\citeauthoryear{Wang and Klabjan}{2018}]{pmlr-v80-wang18d}
Wang, X., and Klabjan, D.
\newblock 2018.
\newblock Competitive multi-agent inverse reinforcement learning with
  sub-optimal demonstrations.
\newblock In {\em Proc. ICML}.

\bibitem[\protect\citeauthoryear{Wang and Sang}{2005}]{Wang2005Multi}
Wang, Y., and Sang, D.
\newblock 2005.
\newblock {\em Multi-agent framework for third party logistics in E-commerce}.
\newblock Pergamon Press, Inc.

\bibitem[\protect\citeauthoryear{Yang \bgroup et al\mbox.\egroup
  }{2018}]{pmlr-v80-yang18d}
Yang, Y.; Luo, R.; Li, M.; Zhou, M.; Zhang, W.; and Wang, J.
\newblock 2018.
\newblock Mean field multi-agent reinforcement learning.
\newblock In {\em Proc. ICML}.

\end{thebibliography}
}
\end{document}